\documentclass{llncs}

\usepackage{makeidx}  

\usepackage{cite}
\usepackage{amsmath,amssymb,amsfonts}
\usepackage{algorithmic}
\usepackage{graphicx}
\usepackage{textcomp}
\usepackage{url}
\usepackage{balance}
\usepackage{enumitem}
\usepackage{xcolor}
\usepackage{verbatim} 
\usepackage{soul}
\usepackage{multirow}
\usepackage{subcaption}
\usepackage{fancyhdr}

\let\oldmaketitle\maketitle
 
\renewcommand{\maketitle}{%
  \oldmaketitle
  \thispagestyle{fancy}
}

\begin{document}
\frontmatter          
\pagestyle{headings}  
\addtocmark{} 
\mainmatter              
\title{Multi-Spectral Image Synthesis for Crop/Weed Segmentation in Precision Farming}
\titlerunning{Data Augmentation for Crop/Weed Segmentation}  
%
\author{Mulham Fawakherji\inst{1} \and Ciro Potena\inst{2}
\and
Alberto Pretto\inst{3} \and\\ Domenico D. Bloisi\inst{4} \and Daniele Nardi\inst{1}}
\authorrunning{Fawakherji et al.} 
%
\tocauthor{Mulham Fawakherji, Ciro Potena, Alberto Pretto, Domenico D. Bloisi, and Daniele Nardi}
\institute{Department of Computer, Control, and Management Engineering,\\ Sapienza University of Rome, Rome, Italy\\
\email{\{fawakherji,nardi\}@diag.uniroma1.it}
\and
Engineering Department, Roma Tre University, Rome, Italy\\
\email{cpotena@os.uniroma3.it}
\and
Department of Information Engineering, University of Padua, Padua, Italy\\
\email{alberto.pretto@dei.unipd.it}
\and
Department of Mathematics, Computer Science, and Economics\\
University of Basilicata, Potenza, Italy\\
\email{domenico.bloisi@unibas.it}
}
\maketitle              

\begin{abstract}
An effective perception system is a fundamental component for farming robots, as it enables them to properly perceive the surrounding environment and to carry out targeted operations. The most recent methods make use of state-of-the-art machine learning techniques to learn a valid model for the target task. However, those techniques need a large amount of labeled data for training. A recent approach to deal with this issue is data augmentation through Generative Adversarial Networks (GANs), where entire synthetic scenes are added to the training data, thus enlarging and diversifying their informative content.
In this work, we propose an alternative solution with respect to the common data augmentation methods, applying it to the fundamental problem of crop/weed segmentation in precision farming. Starting from real images, we create semi-artificial samples by replacing the most relevant object classes (i.e., crop and weeds) with their synthesized counterparts. To do that, we employ a conditional GAN (cGAN), where the generative model is trained by conditioning the shape of the generated object. Moreover, in addition to RGB data, we take into account also near-infrared (NIR) information, generating four channel multi-spectral synthetic images.
Quantitative experiments, carried out on three publicly available datasets, show that (i) our model is capable of generating realistic multi-spectral images of plants and (ii) the usage of such synthetic images in the training process improves the segmentation performance of state-of-the-art semantic segmentation convolutional networks.
\end{abstract}

\section{Introduction} \label{sec:introduction}
\begin{figure}[!t]
    \centering
    \includegraphics[width=.65\columnwidth]{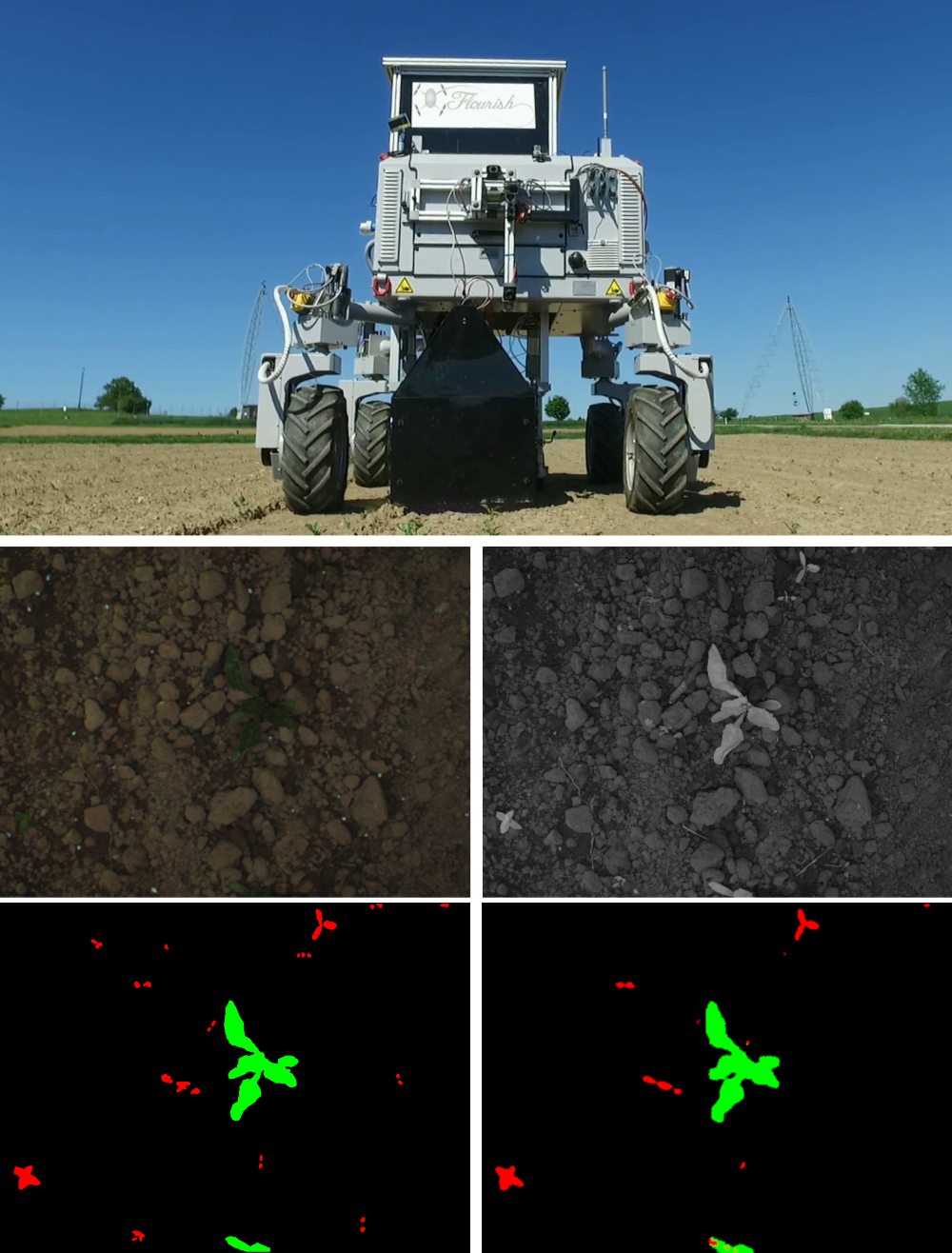}
    \caption{Top row: The BOSCH Bonirob farming robot used to collect the datasets considered in the experiments. The Bonirob is equipped with a Weed Intervention Module (the black structure between the wheels in the picture). This module consists of a perception system for weed classification and a multi-modal actuation systems for weeds removal. Middle row: From left to right, synthetic RGB and synthetic NIR samples, respectively. Bottom row: From left to right, the pixel-wise ground truth and the result obtained by using a semantic segmentation deep neural network, respectively.}
    \label{fig:teaser_img}
\end{figure}

Precision agriculture is a farming management concept based on observing, measuring, and responding to inter and intra-field variability in crops \cite{LEE20102}. A key objective in precision agriculture is the minimization of environmental impacts by reducing the reliance on chemicals products such as herbicides or pesticides.
Farming robots (e.g., see Fig. \ref{fig:teaser_img}, top row) can play an important role in this mission, as they can perform precise weed control through selective treatment applications (e.g., \cite{prettoRAM2020}).

A fundamental requirement to perform selective treatments through robots is to build an effective perception module capable of identifying and localizing crop and weeds in the field and thus 
trigger the proper weeding actions. 
The most commonly adopted approaches use image processing to tackle this problem and rely on machine learning methods, such as Convolutional Neural Networks (CNNs)
\mbox{\cite{potena2016,lottesJFR2016,weednet}}. These data-driven methods allow to train powerful visual classifiers that report high classification performance. However, their performance strongly depends on the size and variety of the training dataset~\cite{xie2016}. This problem is well-known and has been addressed in many different ways (see, among others, \cite{potena2016,dicicco2017}). More recent approaches address this problem by leveraging Generative Adversarial Networks (GANs)~\mbox{\cite{valerio2017arigan,sixt2018rendergan}}. These methods allow to train, in an unsupervised manner, powerful generative models capable of synthesizing photo-realistic images that can be used to increase and diversify the original training datasets. This results in an improved generalization capability of the learned visual classifiers.

In this work, we address the crop and weeds detection task in terms of a semantic image segmentation system  capable of identifying the crop in real images at pixel level, distinguishing it from weeds. Such semantic segmentation task can be effectively tackled by using state-of-the-art data driven, deep learning-based methods such as \cite{Unet,milioto2019icra}. The main disadvantage of such approaches is the need to access large amounts of data provided with accurate pixel-wise semantic annotations (e.g., see the bottom left picture in Fig.~\ref{fig:teaser_img}). A way to automatically generate at least part of this data is desirable. In this context, we propose a novel methodology to synthesize photo-realistic images by using a generative adversarial method. 
Unlike the conventional uses of GANs, which aims to train a model to generate an entire scene, we generate semi-artificial images by replacing only the regions of the scene corresponding to the objects that are relevant to the target perception task (crop and weed plants in our case) with synthesized, photo-realistic counterparts.
The intuition behind this idea is that, usually, vision-based learned classifiers are not able to equally generalize across all the target classes, which in turn can lead to unbalanced classification performance.
To achieve our goal, we use a conditional GAN (cGAN), where the generative model is trained by conditioning the shape of the generated object. This allows to synthesize new realistic objects while keeping their original footprint onto the image, since the generative model receives as an input constraint the object shapes extracted from real objects. 

The main contributions of this work are three-fold. First, we use a cGAN to learn only the data distribution associated to a subset of the target classes, allowing to train more compact generative models and to create photo-realistic training samples in a faster and more effective way.
Second, we perform a quantitative study on our cGAN that estimates the amount of real data needed to generate consistent results.
Third, we use NIR information in order to generate four channel multi-spectral synthetic images.

Using the NIR channel helps to improve accuracy in activities that require vegetation detection. Due to the photosynthesis, healthy green plants absorb more solar energy in the visible spectrum, causing a low reflectance level in the RGB channels. Similarly, the reflectance of the near-infrared spectrum is affected by the same phenomena with opposite results, with a high reflectance level in the NIR channel, where generally 10\% or less of radiation is absorbed \cite{Ustin2020}.

As a further contribution, we created and
made publicly available with this paper a new pixel-wise labeled dataset, the \emph{Sunflower Dataset}, which contains a large number of multi-spectral annotated images acquired over different growing stages in a sunflowers field. The pixel-wise labels highlight the three classes: crop, weed and soil. The Sunflower Dataset and the project’s code are available at:

\url{https://bit.ly/3hHenpE}

To evaluate the effectiveness of the proposed architecture, we report experiments on three publicly available farming datasets, showing that our model is capable of generating realistic 512$\times$512 multi-spectral images of plants (see for instance the middle row of Fig. \ref{fig:teaser_img}), and that the usage of these synthetic images during the training process improves the segmentation performance of state-of-the-art 
semantic segmentation deep neural networks (SSNs). 


The remainder of the paper is organized as follows. After discussing related work in Section \ref{sec:relatedwork}, a brief description of GANs and cGANs is given in Section \ref{sec:preliminaries}.
The proposed method is presented in Section \ref{sec:methods}, while
experimental results are shown in Section \ref{sec:results}. Finally, conclusions are drawn in Section \ref{sec:conclusions}.

\section{Related Work} \label{sec:relatedwork}

A robust crop/weed classification module is an essential component for autonomous farming robots, as it enables the platform to properly perceive the environment and to carry out an efficient weed control policy. The problem has been extensively investigated over the last years and the proposed approaches can be roughly split in two main categories:
\begin{enumerate}
    \item  Classifiers based on hand-crafted features. 
    \item Classifiers based on learned features.
\end{enumerate}    

The methods of the first group usually have limited generalization capabilities, depending on the choice of the features to process.
The approaches in the second category have better generalization capabilities, at the cost of annotating large datasets of images, which is a tedious and time-consuming process. In this section, we focus on crop/weed approaches belonging to the two categories mentioned above. Moreover, we provide a discussion about methods that address the dataset annotation issue.

\subsection{Classifiers Based on Hand-crafted Features}
Methods in this class aim at finding a suitable set of features that have good discrimination properties among the target plant classes.
Haug \textit{et al.}~\cite{haug2014plant} propose a plant classification method that is capable of distinguishing carrots and weeds by using RGB and NIR images. The reported accuracy is around $93.8\%$. Lottes \textit{et al.}~\cite{lottesJFR2016} propose a sugar beets and weeds classification system based on a multi-spectral camera mounted on the robot. The method performs, in sequence, a vegetation detection, an object-based features extraction, a random forest classification, and a smoothing post-process through a Markov random field. Experiments have been carried out in different sugar beets fields reporting good classification performance. This method has been extended in \cite{lottes2017icra}, where the crop/weed classification data are acquired using a camera mounted on a light-weight UAV. The system has been tested in two farms located in Germany and Switzerland, showing good generalization properties and the ability to classify individual plants.
Despite the positive results, the methods based on hand-crafted features are strictly dependent on the choice of the features, which limits their generalization capabilities.


\subsection{Classifiers Based on Learned Features}
Machine learning methods, and more specifically CNNs, offer the potential to overcome the inflexibility of handcrafted vision pipelines, by allowing to develop effective end-to-end classification methods. In this regards, CNNs are usually applied in a pixel-wise fashion, operating on image patches, provided by a sliding window approach. Following this idea, Potena \textit{et al.}~\cite{potena2016} propose a crop/weed classification
architecture composed of a cascade of CNNs. The first CNN detects the vegetation, which is successively used as the input for a second, deeper, CNN that classifies vegetation pixels into crop or weeds.
McCool \textit{et al.}~\cite{mccool2017mixtures} propose a three stage approach. They start from a pre-trained CNN model with state-of-the-art performance, but a high computational cost. Then, a model compression is performed, leading to a faster CNN. Finally, they combine several lightweight models into a mixture model to enhance the performance. They report an accuracy around $93.9\%$.
Mortensen \textit{et al.}~\cite{mortensen2016semantic} use a CNN to estimate the in-field biomass and crop composition. Their method is a modified version of the well-known VGG-16 deep neural network. The reported accuracy is $79\%$ with seven classes of objects.

Differently from classification CNNs, semantic segmentation deep neural networks (SSNs) take images of arbitrary size as input and produce segmented output of corresponding size, without relying on local patches.

Among the many SSNs proposed in the literature, one of the most commonly adopted in crop/weed segmentation is SegNet \cite{Segnet}, which is a deep encoder-decoder architecture for multi-class pixelwise segmentation. Di Cicco \textit{et al.}~\cite{dicicco2017} trained SegNet with real and synthetic images reporting good segmentation performance. Sa \textit{et al.}~\cite{weednet} use SegNet for dense semantic weed classification with multispectral images collected by a Micro Aerial Vehicle (MAV). A similar encoder-decoder architecture is exploited by Milioto \textit{et al.}~\cite{miliotoicra2018}. They augment the RGB input image with task-relevant background knowledge to speed up the training and to better generalize to new crop fields. We exploit a similar idea in our previous work~\cite{FawakherjiYBPN19}, where we propose a pipeline with multiple data channels to support the input of a CNN by using more vegetation indices. These additional information aid the CNN to achieve a good generalization to different crop types.
Lottes \textit{et al.}~\cite{lottes2018icra} propose a crop/weed classification system that, in addition to a Fully Convolutional Network (FCN)~\cite{LongFCN2015}, also exploits the crop arrangement information that is observable from the image sequences. This increases the segmentation performance and the generalization capabilities of the net to previously unseen fields under varying environmental conditions.

\subsection{Labeling Effort Reduction}
The major drawback of both CNNs and SSNs based
architectures is that the level of expressiveness is limited by the size of the training dataset. In the context of precision farming, collecting large annotated datasets involves a significant effort. Datasets should be acquired across different growth stages and weather conditions. Moreover, in the case of SSNs, the pixel-wise annotation process is tedious and extremely time consuming. As a matter of fact, the size of pixel-wise annotated datasets is usually relatively small \cite{xie2016}.

To cope with the above discussed problems, different solutions have been proposed in the literature. Potena \textit{et al.}~\cite{potena2016} propose a novel dataset summarization technique. The main idea is to condense an original, unlabeled, dataset by taking only the most informative images. The summarized dataset will thus lead to a reduced labeling effort while keeping a sufficient segmentation performance. Di Cicco \textit{et al.}~\cite{dicicco2017} use a state-of-the-art graphic engine to generate synthetic and realistic farming scenes. The generated scene, together with the corresponding ground truth data, are used to train the final CNN or to supplement an existing real dataset.
Milioto \textit{et al.}~\cite{miliotoicra2018} propose a CNN that requires little data to adapt to the new, unseen environment. The reported results show a segmentation accuracy around $96\%$ and a fast re-adaptation to the new environments. Sa \textit{et al.}~\cite{weednet} deal with the labeling effort by exploiting different fields with varying herbicide levels, resulting in field patches containing only either crop or weed. This enables to exploit a simple vegetation index as a feature for automatic ground truth generation.
Although the methods described so far can successfully reduce the annotation effort, they may not yet achieve the segmentation performance of a fully trained SSN.

More recent approaches make use of GANs. Giuffrida \textit{et al.}~\cite{valerio2017arigan} exploit a conditional GAN to generate 128$\times$128 synthetic \textit{Arabidopsis} plants, with the possibility to decide the desired number of leaves for the final plant. Their method has been tested using a leaf counting algorithm in order to show how the addition of synthetic data helps to avoid overfitting and to improve the accuracy. Madsen \textit{et al.}~\cite{MADSEN2019147} leverage a GAN to generate artificial image samples of plant seedlings to mitigate for the lack of training data. Their method is capable of generating nine distinct plant species, while increasing the overall recognition accuracy.

As a difference with respect to the last discussed methods, in this work we propose to generate \emph{multi-spectral}
views of agricultural scenes by synthesizing only the objects that are relevant for semantic segmentation purposes. Our method starts by generating a synthetic plant using a cGAN, then the real plant in the image is replaced by a synthetic generated one in order to create a new, semi-artificial image.


\section{Preliminaries} \label{sec:preliminaries}

\begin{figure}[t]
    \centering
    \includegraphics[width=0.9\linewidth]{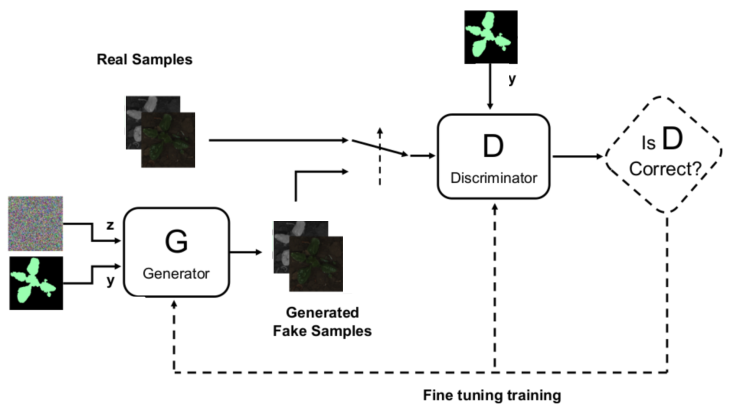}
    \caption{The cGAN generator learns a nonlinear function $G$ that maps an input mask to a photo-realistic image. The cGAN discriminator learns a function $D$ that discerns real from synthesized images produced by $G$.}
    \label{fig:cgan_scheme}
\end{figure}

\subsection{Generative Adversarial Networks}

Generative Adversarial Networks (GANs) \cite{goodfellow2014generative} can estimate generative models through an adversarial training, which simultaneously trains two networks:
\begin{enumerate}
    \item A generative model $G$, whose objective is to capture a data distribution.
    \item A discriminative model $D$ that outputs a single scalar. The goal for $D$ is to estimate the probability that a sample is actually a real data and not a sample synthetically generated by $G$. 
\end{enumerate} 

Given any data distribution $p_{data}(x)$ and a prior input noise distribution $p_z(z)$ (which is typically uniform or Gaussian), the mapping to the data space is represented by $G(z)$, where $G$ is the generative model with its distribution $p_g$. Let us also define the discriminator $D$ as a function that outputs a single scalar $D(x)$ representing the probability that $x$ comes from the real data space rather than from $p_g$.

The training process is carried out by maximizing the probability of $D$ to assign a correct label to the generated and to the real samples, while $G$ is trained to learn the distribution $p_g$ over the data space $x$ so that $D$ can hardly assign them the correct label. $G$ and $D$ are trained in an unsupervised manner by a two-player min-max game that is given by:
\begin{equation}
    \min_{G} \max_{D} V(D,G) = \mathbb{E}_{x \sim p_{data}(x)}[logD(x)] + \mathbb{E}_{z \sim p_z(z)}[log(1-D(G(z)))]
\end{equation}
where $\mathbb{E}$ and $log$ are the expectation and logarithmic operators, respectively. $D$ and $G$ are trained simultaneously until they cannot both improve because $p_g = p_{data}$ and the discriminator is unable to distinguish between the two distributions, i.e., $D(x) = 1/2$.  

\subsection{Conditional Generative Adversarial Networks}

Conditional GANs (cGANs) \cite{Mirza2014cGANs} extend the GAN concept by conditioning both $D$ and $G$ through extra data $y$. The cGAN scheme is shown in Fig. \ref{fig:cgan_scheme}.
It is worth noticing that $y$ can represent any kind of auxiliary information (in \cite{valerio2017arigan}, for instance, $y$ represents the number of leaves of the synthesized plant)
and it is fed into both the generator $G$ and discriminator $D$ as an additional input layer. In the cGAN scheme, $G$ attempts to synthesize realistic images (i.e., fake samples) from the $y$ domain, while $D$ receives samples from both $x$ and $y$ domains and attempts to discern between $(real,real)$ and $(fake,real)$ image pairs.

The loss function of a cGAN can be expressed as:
\begin{equation}
    \min_{G} \max_{D} V(D,G) = \mathbb{E}_{x \sim p_{data}(x)}[logD(x,y)] + \mathbb{E}_{z \sim p_z(z)}[log(1-D(G(z,y)))]
\end{equation}

\section{Proposed Method} \label{sec:methods}

Our goal is to develop an algorithm capable of synthesizing realistic agricultural scenes. Let us define our data distribution $p_{data}(x)$ as a set of images collected by a moving robot in a cultivated field. The images are acquired by a multi-spectral camera that collects NIR images in addition to, and registered with, RGB images. The dataset is annotated in a pixel-wise manner and, for each image, we have a corresponding \textit{total mask} containing information about crop, weed, and soil pixels.

To synthesize new realistic annotated images we need to accomplish two main tasks. 
The first task consists in extracting a \textit{plant mask} from the total image mask. A plant mask is a binary image where the plant pixels that we want to learn are set to 1 and everything else is set to 0.  
The second task consists in learning a function $G:z,y \rightarrow x$ that maps the plant mask $y$ in input to a realistic multi-spectral image. The mapping function $G$ is implemented in a cGAN that contains an implicit model of the conditional probability distribution $p(x|y)$ learned by training. The resulting images will be used as data augmentation to train a deep learning model for crop/weed segmentation. Solving the tasks discussed above has two advantages: 1) it permits to enlarge the training dataset and 2) it allows to diversify the data, thus significantly improving the generalization of the learned models. 

The usage of cGAN as data augmentation tool is not novel and it has been explored in different fields, ranging from medical images, anomaly detection, image classification, and even in the decoding of the position, orientation, and binary ID of markers \cite{valerio2017arigan, 10.3389/frobt.2018.00066,8363678}.
The data augmentation problem is usually addressed by training a generative model capable to reproduce an entire scene, which requires deep models, a large amount of training data, and high computational power.
However, a full scene generation is redundant for our crop/weed segmentation task, where plants are represented by a small percentage of the whole image pixels and the majority of pixels belongs to the background (soil).

Moreover, since the accuracy of a SSN can often vary significantly across classes, in our scenario we can augment the number of training samples only for the classes with a lower classification accuracy. In fact, while the network is able to accurately detect the soil, it usually miss-classifies the pixels belonging to crop and weeds (due to their similar visual appearance). 
In such a case, there is no need to increase the number of soil samples, while increasing the crop samples can provide a significant information gain.

In this work, rather than training a generator $G$ capable to synthesize an entire scene, our idea is to focus on generating instances of some specific object classes, specifically the ones with the lowest segmentation accuracy. In the rest of this section, we describe the three major steps involved in the generation of realistic agricultural samples:

\begin{enumerate}
\item The training of the cGAN for learning the generative model. 
\item The quantitative evaluation of the cGAN training results. 
\item The composition of the synthetic farming scenes.
\end{enumerate}


 


\subsection{cGAN Architecture}
\label{sec:cgan}

The first step of our approach concerns the generation of photo-realistic images of specific classes of objects that populate an agricultural scene.
We employ here the SPatially-Adaptive DEnormalization (SPADE) cGAN architecture~\cite{spade}. Differently from other common cGANs, this type of network performs a semantic image synthesis by converting a semantic segmentation mask into a photo-realistic image. In other words, its input/output behavior is the opposite of an image segmentation network. 

\begin{figure}[t]
\centering
\begin{subfigure}{.5\textwidth}
  \centering
  \includegraphics[width=.7\linewidth]{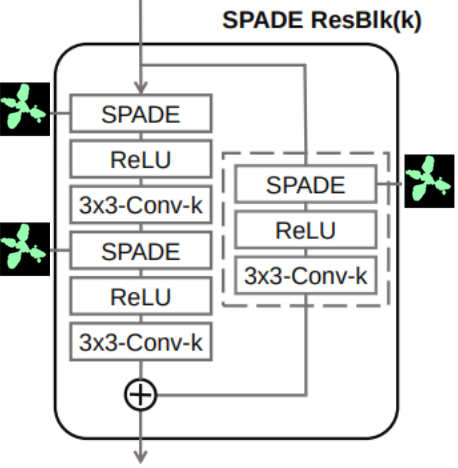}
  \caption{SPADE ResBlk}
\end{subfigure}%
\begin{subfigure}{.5\textwidth}
  \centering
  \includegraphics[width=.7\linewidth]{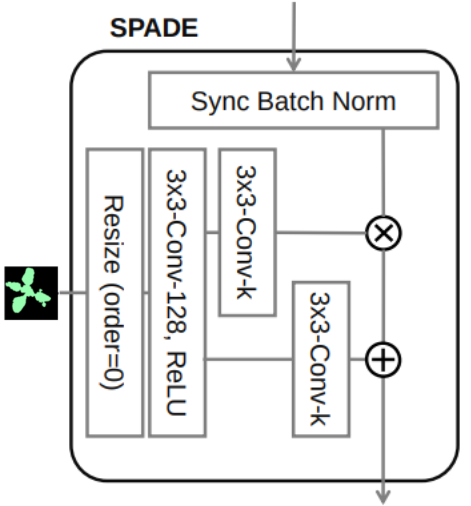}
  \caption{SPADE element}
\end{subfigure}
\caption{SPADE Architecture. (a) SPADE ResBlk. (b) SPADE element. The term 3x3-Conv-k denotes a 3-by-3 convolutional layer with k convolutional filters. The
segmentation map is resized to match the resolution of the corresponding feature map using nearest-neighbor downsampling.}
\label{fig:SPADE_ResBlk}
\end{figure}

In the SPADE architecture, the image encoder encodes a real image into a latent representation for generating a mean and a variance vector. 
This architecture aims to sample from the learned model new realizations, modulating the style in terms of color and texture of the elements of interest, while unchanged keeping the original shapes used in conditioning.
The generator in the SPADE architecture contains a series of basic components, called residual blocks (ResBlk). A SPADE ResBlk (shown in Fig. \ref{fig:SPADE_ResBlk}a) includes some SPADE elements (see Fig. \ref{fig:SPADE_ResBlk}b).
The SPADE generator is built based on the pix2pixHD framework~\cite{wang2018high}. It starts with random noise in the input and uses the semantic map at every SPADE ResBlk layer. 
Using SPADE, it is also possible:
\begin{itemize}
    \item to separate between semantic and style control;
    \item to change the final content, by modifying the semantic map;
    \item to change the style of the image, by modifying the random vector.
\end{itemize}
The SPADE discriminator takes in input a concatenation of the segmentation map with the original (or generated) image and decides if that image is real or fake.
Also the discriminator architecture follows pix2pixHD and it uses a multi-scale design with instance normalization, with the difference that spectral normalization is applied to all the convolutional layers of the discriminator. The encoder is composed of six convolutional layers with stride 2 followed by two linear layers. It is responsible for producing the mean ($\mu$) and covariance ($\sigma^2$). 

Compared to its original version, we have made two major changes to the SPADE architecture:
\begin{enumerate}
    \item The first change is in the SPADE image synthesis modalities. The original version takes as input RGB images and generates RGB images as well. In our case, to exploit the NIR channel, we enabled the network to work with four channel images and thus to generate multi-spectral images.
    \item The second modification has been made to increase the size of the generated samples. The original version generates images with a resolution of 256$\times$256, which may not be enough to generate all the possible object classes. Differently, we generate images with a resolution of 512$\times$512.
\end{enumerate}


\subsection{cGAN Evaluation Metrics}
\label{sec:gan_train_ev}

Evaluating GANs is a very challenging task and several aspects need to be taken into consideration when defining metrics that can produce meaningful scores. These metrics should be capable to distinguish between generated and real samples, to detect overfitting, and to identify mode collapse and mode drop. Our goal is to check whether the learned generative model generalizes well with respect to the problem of photo-realistic crop/weed generation.

For most of the GANs presented in the literature, network inspection is qualitative only, based on a manual inspection to verify the fidelity of the generated sample. This kind of evaluation is still considered the best approach, but it is time-consuming, subjective, and often it can also be misleading.

In this paper, we employ an empirical evaluation based on quantitative metrics. The key idea is to use samples generated by the network and samples collected from the real dataset to extract features from both of them, and then to calculate performance using specific metrics. In particular, we employ six metrics: Inception Score, Mode Score, Kernel MMD, Wasserstein distance, Fréchet Inception Distande (FID) and 1-nearest neighbor (1-NN). For space constraints, we describe here only the Inception Score, being the most popular metric for evaluating GANs, while the definition of all other metrics can be found in~\cite{Xu2018}.

\subsubsection{Inception Score}
It is a metric capable of measuring not only the quality, but also the diversity of generated images using an external model, the Google Inception network \cite{inception}, trained on the ImageNet dataset \cite{imagenet}.
The Inception Score (IS) can be calculated using the following equation:
\begin{equation}
    \textrm{IS}(p_g)=e^{\mathbb{E}_{x \sim p_g}[KL(p_\mathcal{M}(l\mid\mathrm{x}) \parallel p_\mathcal{M}(l))]}
    \label{eq:inception}
\end{equation}

By considering a pre-trained model $\mathcal{M}$, $p_\mathcal{M}(l\mid\mathrm{x})$ refers to the label distribution of $\mathrm{x}$ predicted by $\mathcal{M}$, and $p_\mathcal{M}(l)=\int_{\mathrm{x}}p_\mathcal{M}(l\mid\mathrm{x})dp_g$, which gives the marginal of $p_\mathcal{M}(l\mid\mathrm{x})$ over the probability measure $p_g$. The expectation and the integral in $p_\mathcal{M}(l\mid\mathrm{x})$ can be approximated with independent and identically distributed samples from  $p_g$. The $KL$ operator represents the Kullback–Leibler divergence between the distributions $p_\mathcal{M}(l\mid\mathrm{x})$ and $p_\mathcal{M}(l)$.\\

We used the six metrics listed above as evaluation metrics to give a final quantitative intuition of how much the generated fake samples are close to the real data distribution. First, we identify a reference value by computing the metrics over two sets of samples from the real dataset. This process is repeated ten times with random selection at each time and getting the mean. Then, for each cGAN model output, we generate a set of fake samples and compute the metrics between the generated fake samples and the real data.

\subsection{Agricultural Scene Composition}
\label{sec:scene_gen}

\begin{figure}[!h]
 \centering
 \includegraphics[width=\linewidth]{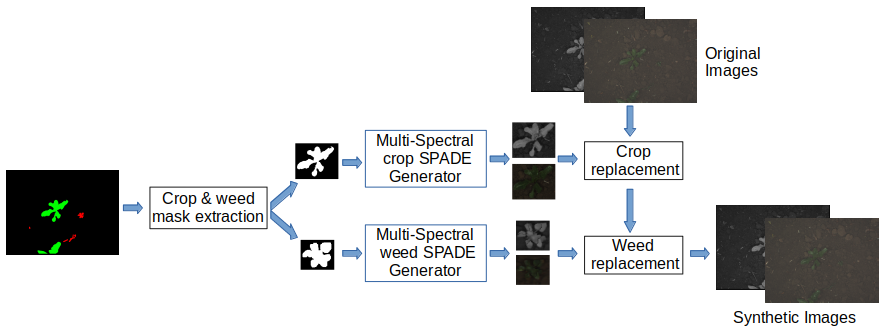}
 \caption{
 Dataset creation for segmentation training. First the crop and weed masks are taken from the full mask, then new RGB+NIR crops and weeds are generated from these masks and pasted back into the original image.}
 \label{fig:segmentation_dataset}
\end{figure}

\begin{figure}[!h]
    \centering
    \includegraphics[width=0.65\columnwidth]{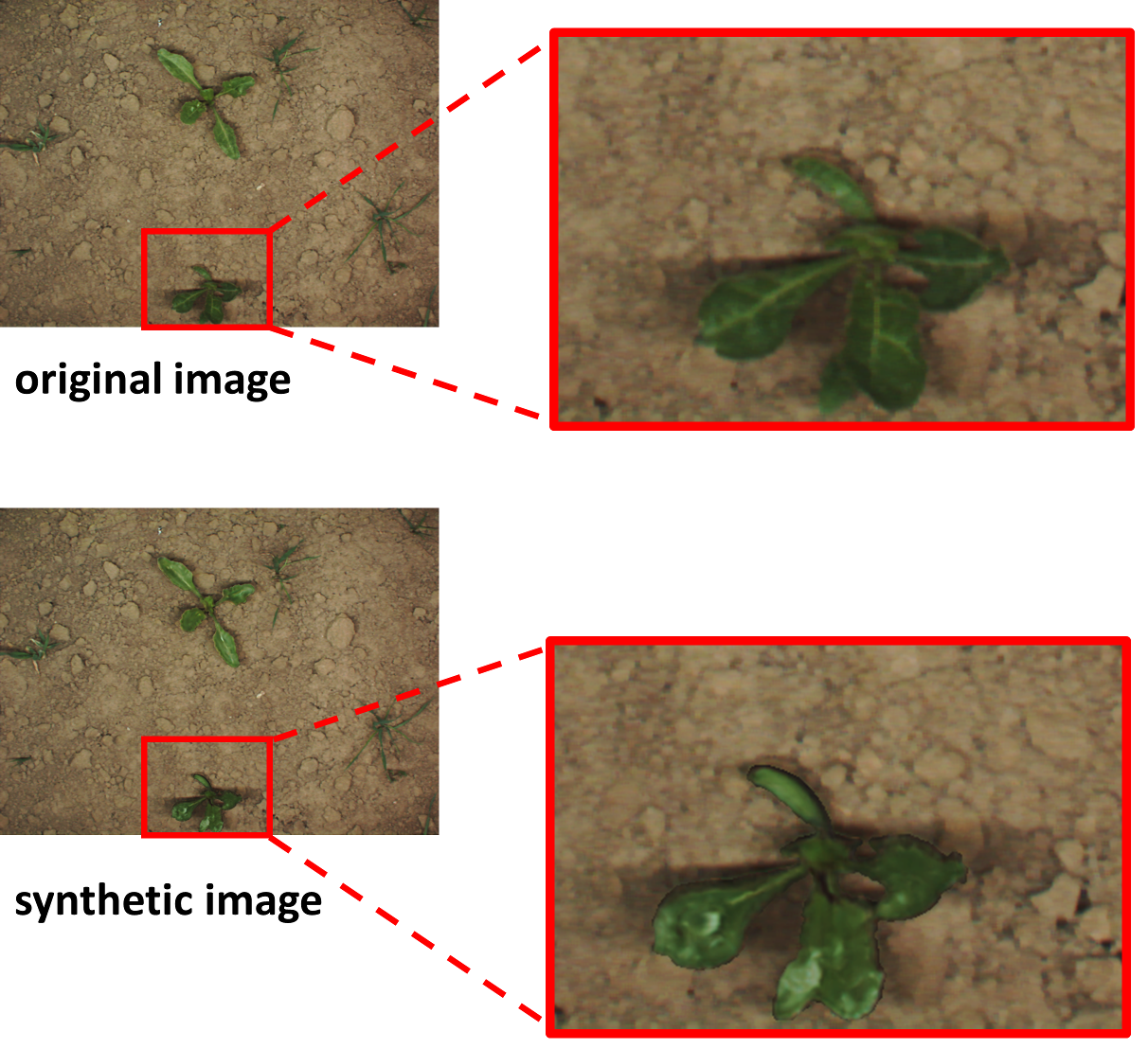}
    \caption{An original and a synthetic image. The synthetic image is obtained by inserting in the original image a plant sample generated by using our cGAN.}
    \label{fig:detail}
\end{figure}

In the final step, our approach uses the crop and weeds RGB/NIR images described in Section~\ref{sec:cgan} to create a realistic agricultural scene. To do so, we follow the scheme shown in Fig.~\ref{fig:segmentation_dataset}.

First, we get the crop and weed masks from the total image mask. The mask is then resized to the SPADE network input size, which is 512$\times$512 pixels for crop and 128$\times$128 for weeds. 
This difference is due to the fact that weeds tend to be smaller than the crop. The SPADE network then generates a random, photo-realistic, crop/weeds instance by using the shape of the input mask and a random noise signal. The generated image is then replaced into the source image.
Fig.~\ref{fig:detail} shows an example of the obtained synthetic image. 

\section{Experimental Results} \label{sec:results}

A quantitative evaluation has been carried out to show that by augmenting the training datasets with synthetic photo-realistic images generated with our model it is possible to:
\begin{enumerate}
    \item Improve the generalization capability of the chosen segmentation network;
    \item Increase the performance in crop/weed segmentation.
\end{enumerate}
Moreover, the annotation effort is reduced, since the cGAN generates both the images and the masks.
We also performed a quantitative study to test the amount of real data needed to train a SPADE cGAN model capable of generating good plant samples, i.e., synthetic images close enough to the real data.
\begin{figure}[!t]
    \centering
    \includegraphics[width=.7\columnwidth]{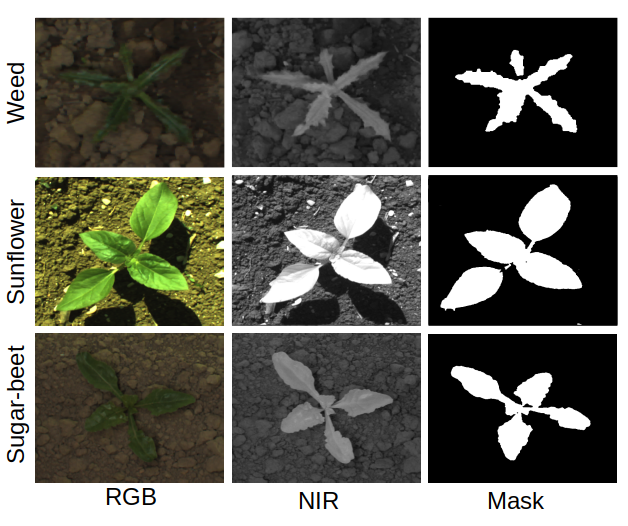}
    \caption{Training data for our SPADE module. First row: examples of data used to train SPADE for weeds. Second row: examples of data used to train SPADE for sunflowers. Third row: examples of data used to train SPADE for sugar beets. The columns show (from left to right) RGB, NIR, and mask samples.}
    \label{fig:cGan_dataset}
\end{figure}


\subsection{Experimental setup}


We performed experiments on three publicly available datasets, considering two different types of crops, namely sugar beet and sunflower.

\subsubsection{Sugar beet datasets.}
For sugar beet, we used two publicly available datasets: the Bonn dataset and the Stuttgart dataset~\cite{chebrolu2017ijrr}. Both datasets have been collected by using a BOSCH Bonirob farm robot moving on a sugar beet field across different weeks. The datasets are composed of images taken by a 1296$\times$966 pixels 4-channels (RGB + NIR) JAI AD-13 camera, mounted on the robot and facing downward. An example of sugar beet taken from Bonn dataset is shown at the bottom of Fig. \ref{fig:cGan_dataset}. 

\subsubsection{Sunflower dataset.}
In this work, we introduce a new dataset for crop/weed segmentation, called Sunflower dataset\footnote{\texttt{http://www.diag.uniroma1.it/{\texttildelow}labrococo/fsd/sunflowerdatasets.html}} that has been collected by the authors of this paper. An example of sunflower taken from the Sunflower dataset is shown in the middle of Fig. \ref{fig:cGan_dataset}. Data has been acquired by using a custom-built agricultural robot moving in a sunflower farm in Jesi, Italy. The dataset has been recorded in spring, across a period of one month, starting from the emergence stage of the crop plants, until the end of the useful period for the use of chemical treatments. As for the Bonn and the Stuttgart datasets, images were acquired using a 1296$\times$966 pixels 4-channels (RGB
+ NIR) JAI AD-13 camera, mounted on the robot and facing downward. The Sunflower dataset, composed of 500 images, provides RGB and NIR images with pixel-wise annotation of 3 classes: crop, weed, and soil. It is organized into three subsets:
\begin{itemize}
\item \textit{Jesi-05-12}, which includes images of sunflower crops in the emergence stage.
\item \textit{Jesi-05-18}, which includes images of sunflower crops in a subsequent growth stage.
\item \textit{Jesi-06-13}, which includes images of sunflower crops few days before the end of the period for using chemical treatments. 
\end{itemize}

\begin{figure}[t]
\centering
\begin{subfigure}{.5\textwidth}
  \centering
  \includegraphics[width=0.9\linewidth]{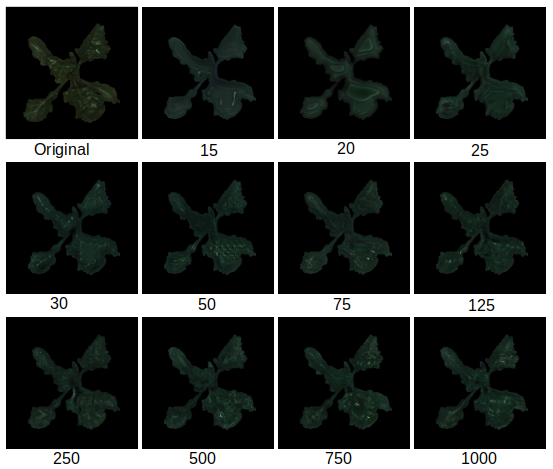}
  \caption{RGB}
\end{subfigure}%
\begin{subfigure}{.5\textwidth}
  \centering
  \includegraphics[width=0.9\linewidth]{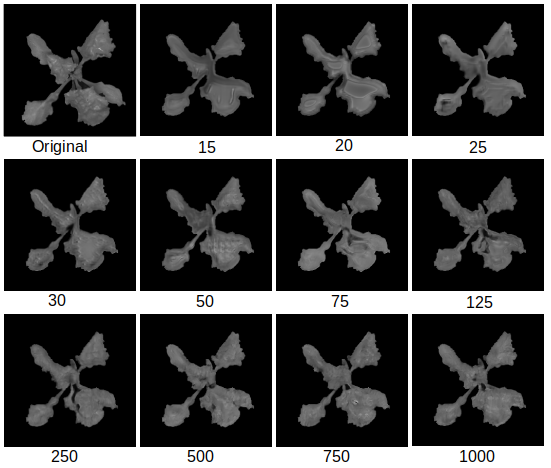}
  \caption{NIR}
\end{subfigure}
\caption{Examples of SPADE models outputs. The number under each image represents the cardinality of the dataset used to train the cGAN. }
\label{fig:Models_output}
\end{figure}

\subsection{cGAN Impact Evaluation}
Our method requires a certain number of labeled real images to effectively train the cGAN network. In this experiment, we measure the minimum number of images needed to train the cGAN in order to achieve a positive impact on the segmentation performance.
Specifically, we trained in turn the SPADE cGAN network by using datasets with different cardinality, respectively with 15, 20, 25, 30, 50, 75, 125, 250, 500, 750, and 1,000 images extracted from the Bonn sugar beet dataset. Then we used each trained model to generate sugar beet crop images.
An example of output image from each model can be seen in Fig. \ref{fig:Models_output}.

We performed two kinds of evaluation for the trained models. In the first one, we used the metrics described in Section \ref{sec:gan_train_ev}. We computed the mean of the metrics over 20 random selected sets from the real dataset. 
We saved these values as a reference for later comparison. Then, we computed the metrics for each model.
Finally, to retrieve a single value representing the best model, we computed the mean error between the trained models metrics and the reference metrics. Table \ref{tab:Generalization_Evaluation} shows the evaluation results. In this table, each model is named with \emph{SPADE-N}, where \emph{N} is replaced with the cardinality of the dataset used to train the model. Generally, the higher the number of images, the smaller the error.

\begin{table}[t]
\centering
\caption{Error over six evaluation metrics between SPADE  models generated samples and real samples.}
\begin{tabular}{l c c c c c c c}
\hline
Model& EMD   & FID & Inception&KNN & MMD & MODE & Mean error \\ \hline
                

\textit{SPADE-15} &21.74  & 0.16  & 0.95 & 0.49 &  0.56  & 1.38 & 4.21   \\  
\textit{SPADE-20} & 21.35 &  0.29 & 0.70 & 0.49 & 0.55 &  1.45 & 4.13 \\  
\textit{SPADE-25} & 17.05  & 0.33  & 0.65 & 0.43 &  0.40  & 1.18 & 3.34   \\   
\textit{SPADE-30} & 9.4 & 0.1 & 1.1 & 0.4 & 0.30 & 1.11 & 2.0  \\   
\textit{SPADE-50} & 10.1 & 0.3 & 1.2 & 0.28 & 0.235 &  \textbf{1.03} & 2.2\\   
\textit{SPADE-75} &7.53 & 0.02 & 1.19 & 0.42 & 0.27 & 1.28& 1.8  \\  
\textit{SPADE-125} &8.73 & 0.23 & \textbf{0.68} & 0.43 & 0.25 & 1.28 & 1.93  \\   
\textit{SPADE-250} & 5.57 & 0.2 & 1.3 & 0.29 & 0.19 & 1.33 & 1.5  \\   
\textit{SPADE-500} & 4.1 & 0.16 & 0.92 & 0.2 & 0.14 & 1.082& 1.1  \\   
\textit{SPADE-750} & 1.04 & \textbf{0.02}  &  1.31 & 0.27 &0.193 & 1.2  & 0,7 \\   

\textit{SPADE-1000} &\textbf{0.03}  & 0.07 & 1.17 & \textbf{0.09} & \textbf{0.07} & 1.15& \textbf{0.43}   \\  \hline

\end{tabular}
\label{tab:Generalization_Evaluation}
\end{table}

\begin{table} [t]
\centering
\caption{Segmentation performance for the Bonnet architecture, trained on two different inputs, namely RGB and RGB + NIR, by using different training sets augmented with a varying amount of synthetic data.}
\begin{tabular}{l c c c c | c c c c}
\cline{1-9}
& \multicolumn{4}{c|}{RGB} & \multicolumn{4}{c}{RGB+NIR} \\
\hline

  Model &  &  \multicolumn{3}{c|}{IoU} &  &  \multicolumn{3}{c}{IoU}   \\ 
 \cline{3-5}\cline{7-9}
& mIoU & Soil  &
     Crop  &
     Weed &mIoU & Soil  &
     Crop   &
     Weed\\ 

\textit{Mix-15} & 0.51 & 0.99 & 0.16& 0.38 & 0.52 & 0.99 &
      0.22 &
      0.36  \\

\textit{Mix-50} &  0.53 &0.99 &
     0.21 &
     0.38 & 0.59  &  0.99 & 0.22& 0.57 \\

\textit{Mix-75}& 0.62 & 0.99& 0.33&0.53 & 0.63& 0.99  & 0.43
  & 0.25  \\

\textit{Mix-125}&0.58 &  0.99 & 0.21& 0.54& 0.72& 0.99 & 0.37& 0.80\\ 

\textit{Mix-500}&0.72 &0.99 &0.35 &0.81 &   0.77 & 0.99 &
      0.50 &
      0.83   \\  
\textit{Mix-750}&0.73 & 0.99 & \textbf{0.41}& 0.81 &  0.81& 0.99 &
      0.53 &
      0.90   \\

\textit{Mix-1000} &\textbf{0.76} &  0.99    &0.38&  \textbf{0.92} &\textbf{0.82} &0.99  & \textbf{0.55}& \textbf{0.92} 
\\
\hline

\end{tabular}
\label{tab:Generalization_Evaluation_2}
\end{table}

In the second evaluation, we used the trained SPADE models to generate different datasets for semantic segmentation with synthetic crop, by using the proposed approach. We then augmented the real dataset with synthetic generated ones and we used such augmented datasets to train Bonnet \cite{milioto2019icra}, which is a state-of-the-art semantic segmentation network for precision farming. Finally, we evaluated each trained model by using part of the real dataset not used in training as test data. The results are shown in Table~\ref{tab:Generalization_Evaluation_2}, where the Intersection over Union (\textit{IoU}) and Mean Intersection over Union (\textit{mIoU}) metrics were used \cite{10.1007/s11263-014-0733-5}. Each model is named \emph{Mix-N}, where \emph{N} is replaced with the cardinality of the dataset used to train the cGAN to generate the synthetic images used in the dataset augmentation. The results show that the IoU increases by increasing the cardinality of the dataset used to train the SPADE network.


\subsection{Semantic Segmentation Results}
We have carried out four experiments to show that, by augmenting the  training  datasets with synthetic photo-realistic images, it is possible to increase the performance in crop/weed segmentation tasks.

\begin{enumerate}
    \item In the first experiment, we augmented the Bonn dataset with synthetic photo-realistic images of sugar beet and weed plants, obtaining better results;
    \item In the second experiment, we show the importance of using multi-spectral images;
    \item In the third experiment, we compared the use of traditional augmentation techniques with our method, demonstrating that the use of cGAN is a winning strategy;
    \item In the fourth experiment, we augmented the sunflower dataset with synthetic photo-realistic images of sunflower plants, obtaining better results.
\end{enumerate}

In the following experiments, we divided the data into different, non-overlapping subsets composed by a set of images for training and a set of images for test.

\subsubsection{Experiment 1: Augmenting the Sugar Beet Dataset.}
 In this experiment, we trained two different types of SPADE networks. The first one was trained to generate sugar beet crops, the second one was trained to generate weeds (general types of weed, similar to those one appearing in the Bonn dataset).
 
 Using data from both the Bonn and Stuttgart sugar beet datasets, we created four different datasets:
\begin{enumerate}
\item \textit{Original}, which is a reduced version of the Bonn dataset. We took a total of 1,600 images from the Bonn dataset, randomly chosen among different days of acquisition in order to contain different growth-stages of the target crop. Then, we split it into a training set (1,000 images), a validation set (300 images), and a test set (300 images). 
\item \textit{Synthetic Crop}, composed of 1,000 images with synthetic crop generated by using our architecture.
\item \textit{Synthetic Weed}, composed of 1,000 images with synthetic weeds generated by using our architecture.
\item \textit{Mixed}, containing 1,000 images and composed by the union of 500 images from the Original dataset and 500 images with synthetic crop and weeds.
\end{enumerate}

For testing, we used 300 real images from the Stuttgart dataset and 300 real images (not used for training) from Bonn dataset. 
It is worth noting that we used the Stuttgart dataset to show the improvement in the generalization capability of the segmentation network after augmenting the training dataset with our approach.
We point out that in the synthetic datasets we replaced with synthetically generated samples only the plants whose stem is totally framed into the image. For the plants that are mostly out of the frame, the original one is kept. We experimentally verified that it is necessary to have the stem of the plant roughly in the center the mask to obtain an effective synthetic image generation.
 
\begin{figure}[!h]
    \centering
    \includegraphics[width=.95\columnwidth]{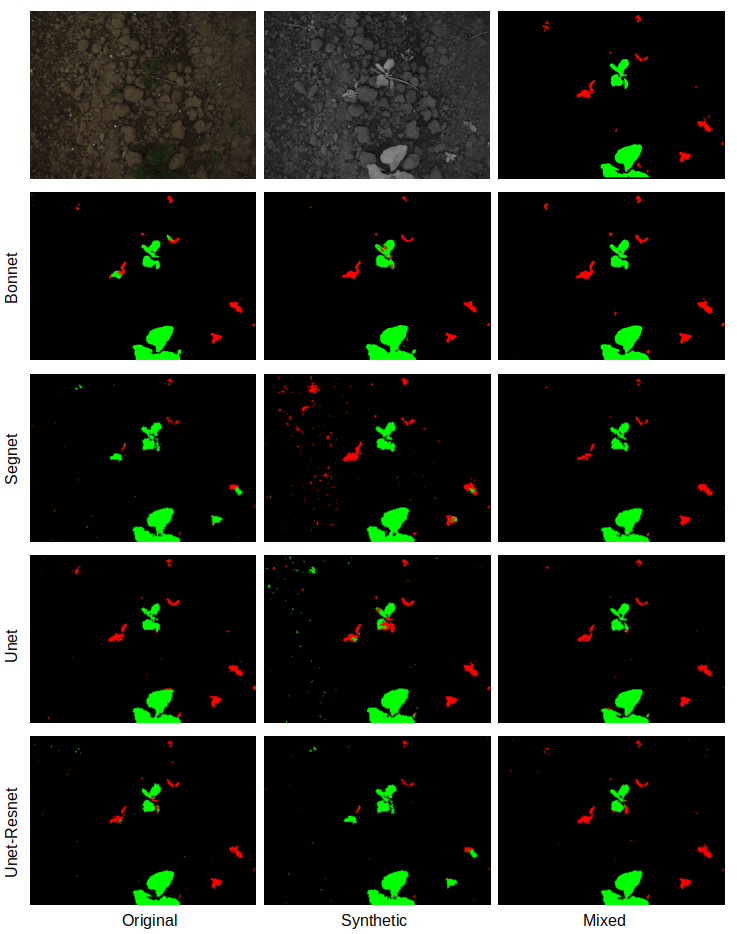}
    \caption{Examples of a segmented image from  Bonn sugar beet test set obtained by using four different segmentation networks trained with three different datasets. The first row of the image shows the input RGB, NIR images and their corresponding ground truth. The remaining rows show the segmentation results generated by the different networks on the $Original$, $Synthetic$, and $Mixed$ training datasets.}
    \label{fig:pred_ex}
\end{figure}

We used the four datasets described above to train four state-of-the-art semantic segmentation networks, namely U-Net \cite{Unet}, UNet-ResNet (U-Net with ResNet50 back-end), Bonnet \cite{milioto2019icra}, and SegNet \cite{Segnet}.
An example of segmentation results is shown in Fig.~\ref{fig:pred_ex}.

\begin{table}[t]
\centering
\caption{Pixel-wise  segmentation performance with RGB input, networks trained on four different datasets, tested on part of Bonn dataset.}
\begin{tabular}{c c c c c c c c c c c c}
 \hline
    SSN   &   Dataset  & mIoU   & \multicolumn{3}{c}{IOU} & \multicolumn{3}{c}{Recall} &\multicolumn{3}{c}{Precision} \\  \cline{4-6} \cline{8-9} \cline{11-12}
            &     &      & Soil   &  Crop  &  Weed & &  Crop  &  Weed & &  Crop  &  Weed\\  \hline

\multirow{4}{*}{\textit{Bonnet}}
     &\textit{Synthetic Weed}      & 0.65 & 0.99 & 0.64 & 0.31  & & 0.71 & 0.53 & & 0.64 & 0.38 \\   
      
     & \textit{Synthetic Crop}     &0.67&  0.99  & 0.73  & 0.30 &    & 0.93  & 0.49   &  & 0.73   & 0.45 \\  
     &\textit{Original} & 0.70 & 0.99 & 0.75 & 0.35 &  & 0.84 & 0.54 & & 0.82 & 0.49 \\
     & \textit{Mixed}    & \textbf{0.76} &  0.99  &  \textbf{0.92}  & \textbf{0.38} &    & \textbf{0.96}  & \textbf{0.58}   & & \textbf{0.95}   & \textbf{0.56 }\\   \hline

\multirow{4}{*}{\textit{UNet-ResNet}}
    &\textit{Synthetic Weed } & 0.64 & 0.99 & 0.73 & 0.18  & & 0.73 & 0.20 &  & 0.86 & 0.30 \\  
    & \textit{Synthetic crop} &   0.69 & 0.99  & 0.79 & 0.29 &   & 0.81  & 0.33   &   & 0.96   & 0.18 \\  
    &\textit{Original}  & 0.67 & 0.99 & 0.81 & 0.23 &  & \textbf{0.95} & 0.25 &  & 0.95 & \textbf{0.63} \\ 
    &\textit{Mixed }   &  \textbf{0.72} & 0.99  & \textbf{0.85}  &\textbf{0.32}  &   & 0.87  & \textbf{0.45}   & & \textbf{0.97}   & 0.48  \\  
\hline

\multirow{4}{*}{\textit{U-Net}}
& \textit{Synthetic Weed} & 0.63 & 0.99 & 0.71 & 0.19 &  & 0.74 & 0.19 &  & 0.93 & 0.35 \\    
&\textit{Synthetic Crop} & 0.65 & 0.99  & 0.76  & 0.20 &   & 0.81  & \textbf{0.37}   &  & 0.98   & 0.37 \\ 
& \textit{Original}     & 0.68 & 0.99 & 0.82 & 0.22 & & 0.84 & 0.26 &  & 0.95 & 0.64 \\  
& \textit{Mixed}    & \textbf{0.71}& 0.99  &\textbf{0.87}  & \textbf{0.27} &   & \textbf{0.89}  & 0.28   & & \textbf{0.97}   & \textbf{0.72} \\   \hline

\multirow{4}{*}{\textit{SegNet}}
&\textit{Synthetic Weed}& 0.61 & 0.99 & 0.60 & 0.14 &  & 0.72 & 0.18 &  & 0.86 & 0.47 \\  
&\textit{Synthetic Crop}  & 0.64& 0.99  & 0.74  & 0.17 &   & 0.80  & 0.21   &  & 0.91   & 0.48  \\   
&\textit{Original}  & 0.66 & 0.99 & 0.78 & 0.20 & & 0.81 & 0.37 & & \textbf{0.97} & 0.37  \\ 
&\textit{Mixed}   &\textbf{0.70} & 0.99  & \textbf{0.85}  &\textbf{0.26}  &   & \textbf{0.88}  & \textbf{0.40}   & & 0.96   & \textbf{0.49} \\  \hline

\end{tabular}
\label{tab:bonnet_segmentation_results}
\end{table}

\begin{table}[t]
\caption{Pixel-wise segmentation performance, networks trained on two different inputs (RGB and RGB + NIR), tested on two different datasets.}
\begin{center}
\begin{tabular}{ c  c  c  c   c| c  c  c| c  c  c }
 \hline
 Train set & Test set 
                     &\multicolumn{3}{c }{Bonnet}
                     &\multicolumn{3}{c}{UNet-ResNet}
                     &\multicolumn{3}{c}{U-Net}
                      \\ 
            &        &     
                     & \multicolumn{2}{c  }{IOU}
            &        & \multicolumn{2}{c  }{IOU}
            &        & \multicolumn{2}{c  }{IOU}    \\ \cline{4-5}\cline{7-8}\cline{10-11} 
            &        & mIoU & Crop & Weed   & mIoU & Crop & Weed    & mIoU & Crop & Weed \\   \hline

   \textit{Original}  
        & \textit{Stuttgart  } 
               & 0.30 & 0.13 & 0.1 
               & 0.31 & 0.11 & 0.12 
               & 0.33 & 0.14 & 0.08 
                  \\
       \scriptsize \textit{(RGB)}      
         & \textit{Bonn}  
               & 0.70 & 0.75 & 0.35
               & 0.67 & 0.81 & 0.23
               & 0.68 & 0.82 & 0.22      \\ \hline
      \textit{Mixed } 
         & \textit{ Stuttgart  }
               & 0.38 & 0.26 & 0.18 
               & 0.35 & 0.2  & 0.13  
               & 0.37 & 0.21 & 0.15
                \\
     \scriptsize \textit{(RGB)}
        & \textit{Bonn}
               & 0.76 & \textbf{0.92} & 0.38
               & 0.72 & 0.85 & 0.32
               & 0.71 & \textbf{0.87} & 0.27
                 \\ \hline
       \textit{Original} 
            & \textit{Stuttgart } 
               & 0.49 & 0.32 & 0.12 
               & 0.47 & 0.30 & 0.15
               & 0.45 & 0.34 & 0.13 
                 \\
        \scriptsize \textit{ (RGB+NIR) } 
              & \textit{Bonn} 
               & 0.77 & 0.85 & 0.45
               & 0.31 & 0.85 & 0.35 
               & 0.69 & 0.84 & 0.24
                \\ \hline

  \textit{Mixed} 
               & \textit{ Stuttgart } 
               &\textbf{ 0.57} & \textbf{0.46} & \textbf{0.28}
               & \textbf{0.54} & \textbf{0.52} & \textbf{0.16}
               & \textbf{0.53} & \textbf{0.50} &\textbf{ 0.19} 
                  \\  
 \scriptsize \textit{(RGB+NIR) }
            & \textit{Bonn } 
               &\textbf{0.80}  & 0.88 &\textbf{0.55}  
               & \textbf{0.77} & \textbf{0.92} & \textbf{0.40}
               & \textbf{0.74} & 0.85 & \textbf{0.37} 
                  \\  \hline

\end{tabular}
\end{center}
\label{tab:Multi_channel_segmentation_results}
\end{table}

To evaluate the semantic segmentation output, we used the following metrics: Per-class Intersection over Union, Mean Intersection over Union (denoted as \textit{mIoU}), Per-class Recall, and Per-class Precision.

Table~\ref{tab:bonnet_segmentation_results} shows the quantitative results of the semantic segmentation on real images held out from Bonn dataset. For all the architectures,  the results show that the IoU increases by using the original dataset augmented with the synthetic ones compared to using only the original dataset. 
Additionally, we can see that the rate of correctly predicted crop and weed samples increased across all architectures when we used the mixed dataset for training. For example, in Bonnet architecture, the correctly predicted samples increased more than $10\%$ in  case of sugar beet, and around $4\%$ for weed samples.
Moreover, using only the synthetic dataset also leads to a competitive performance when compared to using only the original one. In the case of the UNet-ResNet architecture, using only the synthetic dataset overcomes the performance obtained by using only the original one.
We can notice also that crop has a more positive impact on the dataset used in semantic segmentation and that is because the quality of synthetically generated crops are better than the quality of synthetic generated weeds. This behavior stems from the fact that weeds have more diverse types and shapes, which makes the task of cGAN of capturing weed style distribution more challenging.
\subsubsection{Experiment 2: The Importance of Multi-Spectral Images.}
  To show the contribution of having both multi-spectral and synthetic data augmentation, we considered four different training sets, i.e., \textit{Original} and \textit{Mixed} containing RGB images only and \textit{Original} and \textit{Mixed} containing both RGB and NIR images. We used the Stuttgart and Bonn datasets as test data. Table~\ref{tab:Multi_channel_segmentation_results} shows the segmentation results for this experiment. For all the tested architectures, the segmentation capability improves when using the \textit{Mixed} dataset, i.e., when the dataset containing real images is augmented with synthetic data. This supports the idea of creating artificial samples to improve the segmentation performance.

Moreover, the results in Table~\ref{tab:Multi_channel_segmentation_results} show that using the \textit{Mixed} RGB plus NIR dataset during the training process leads to a better performance. In fact, the segmentation performance increases in all the considered setups. 
This proves our claim that also the NIR channel generated using our approach improves the segmentation capability of all the convolutional network architectures used in our experiments.

\subsubsection{Experiment 3: Comparison with Traditional Augmentation Techniques.}

Basic image manipulations to obtain training data augmentation include rotation, shifting, flipping, zooming, and cropping. Also texture manipulations like Gaussian and median blurring, noise injection, and contrast and brightness variation can be used to augment the available data. The aim of this experiment is to show the effectiveness of our method over the traditional augmentation strategies. We prepared four training datasets as follows.
\begin{itemize}
    \item \textbf{Basic augmentation}: 1,000 original images augmented with 1,000 images using basic image operations. 
    \item \textbf{Texture augmentation}: 1,000 original images augmented with 1,000 images using texture manipulations.
    \item \textbf{Ours + Basic augmentation}: 1,000 original images augmented with 500 images using our strategy plus 500 images as in basic augmentation;
    \item \textbf{Ours + Texture augmentation}: 1,000 original images augmented with 500 images using our method plus 500 images processed with texture manipulation strategies.
\end{itemize}

Results are reported in Table~\ref{tab:augmentation_resulats}. We can see that the best results are achieved by the model trained on the dataset augmented using both our method and a basic augmentation, while augmenting the dataset with our method plus texture manipulations cause a drop in the mIoU.

\begin{table}[t]
\centering
\caption{Segmentation results of Bonnet architecture, trained on four different datasets, tested on Bonn test dataset}
\begin{tabular}{l c c c }
\cline{1-4}

  \multirow{2}{*}{Augmentation Strategy} &  &  \multicolumn{2}{c}{IoU}   \\ 
 \cline{3-4}
& mIoU & Crop  &
     Weed \\ 
\hline
\textit{Basic augmentation} & 0.71 & 0.76 & 0.37 \\ 
\textit{Texture augmentation} & 0.73 & 0.79 & 0.40 \\ 
\textit{Ours + Basic augmentation} & \textbf{0.78}  & \textbf{0.93} & \textbf{0.43}
      \\

\textit{Ours + Texture manipulation} &  0.66 &0.80&
     0.19   \\   
  
\hline

\end{tabular}
\label{tab:augmentation_resulats}
\end{table}

\begin{table}[t]
\centering
\caption{ Pixel-wise segmentation performance for Sunflower Dataset, networks trained on two different inputs (RGB and RGB + NIR).
}
\begin{tabular}{c c c c c c c c c c c c}
 \hline
      \multirow{2}{*}{Dataset} & \multirow{2}{*}{SSN}    & mIoU   & \multicolumn{3}{c}{IOU} & \multicolumn{3}{c}{Recall} &\multicolumn{3}{c}{Precision} \\  \cline{4-6} \cline{8-9} \cline{11-12}
            &     &      & Soil   &  Crop  &  Weed & &  Crop  &  Weed & &  Crop  &  Weed\\  \hline

\textit{Synthetic}
  &\textit{Bonnet } &0.76&  0.99  & 0.84  & 0.46 &    & 0.94  & 0.61   &  & 0.88   & 0.67 \\ 
\textit{Crop}   &\textit{U-Net } &0.51&  0.98  & 0.038  & 0.51 &    & 0.039  & 0.51   &  & 0.57   & 0.98 \\ 
 \textit{(RGB)}   &\textit{Unet-resnet } &0.53&  0.98  & 0.05  & 0.59 &    & 0.05  & 0.63   &  & 0.32   & 0.89 \\ 
  
\hline
\textit{Synthetic} &\textit{Bonnet} &   0.83 & 0.99  & 0.84 & 0.66 &   & 0.97  & 0.81   &   & 0.85   & 0.78 \\  

\textit{Crop} &\textit{U-Net } &0.701&  0.99  & 0.69  & 0.41 &    & 0.81  & 0.45   &  & 0.83   & 0.87 \\ 
 \textit{(RGB+NIR)}   &\textit{UNet-ResNet } &0.704&  0.99  & 0.65  & 0.47 &    & 0.73  & 0.52   &  & 0.86   & 0.82 \\ 
\hline

\multirow{2}{*}{\textit{Original}}
& \textit{Bonnet}    & 0.70 & 0.99 & 0.82 & 0.30 &  & 0.94 & 0.61 & & 0.86 & 0.23 \\ 
 &\textit{U-Net } &0.39&  0.97  & 0.15  & 0.031 &    & 0.17  & 0.04   &  & 0.72   & 0.10 \\ 
 \textit{(RGB)}   &\textit{UNet-ResNet } &0.43&  0.98  & 0.28  & 0.04 &    & 0.34  & 0.06   &  & 0.65   & 0.08 \\ 
\hline
\multirow{2}{*}{ \textit{Original} }
& \textit{Bonnet}    & 0.80 & 0.99 & 0.78 & 0.62 &  & 0.87 & 0.82 &  & 0.88 & 0.71 \\ 
&\textit{U-Net } &0.64&  0.99  & 0.54  & 0.38 &    & 0.70  & 0.42   &  & 0.83   & 0.40 \\ 
 \textit{(RGB+NIR)}   &\textit{UNet-ResNet } &0.66&  0.99  & 0.60  & 0.43 &    & 0.75  & 0.49   &  & 0.88   & 0.90 \\ 
\hline

\multirow{2}{*}{\textit{Mixed} }
&\textit{Bonnet} & 0.78 &  0.99  & 0.87  & 0.48 &    & 0.95  & 0.59 & & 0.91   & 0.73 \\ 
 &\textit{Unet} &0.59 &  0.98  & 0.64  & 0.12 &    & 0.71  & 0.12 & & 0.88   & 0.98 \\ 
\textit{(RGB)}&\textit{Unet-resnet} &0.60 &  0.98  & 0.67  & 0.15 &    & 0.74  & 0.14 & & 0.90   & \textbf{0.97} \\
\hline

\textit{Mixed}
&\textit{Bonnet} & \textbf{0.86}  & \textbf{0.99}   & \textbf{0.88}  & \textbf{0.69} &   & \textbf{0.97}  & \textbf{0.85}   & & \textbf{0.90}   & 0.79  \\ 
&\textit{U-Net} &  0.69 & 0.99  & 0.68  & 0.40 &   & 0.80  & 0.43   & & 0.83   & 0.87  \\ 
\textit{(RGB+NIR)} &\textit{UNet-ResNet} &  0.72 & 0.99  & 0.70  & 0.48 &   & 0.77  & 0.51   & & 0.88   & 0.87  \\
\hline

\end{tabular}
\label{tab:sunflower_segmentation_results}
\end{table}

\begin{figure}[!h]
    \centering
    \includegraphics[width=.95\columnwidth]{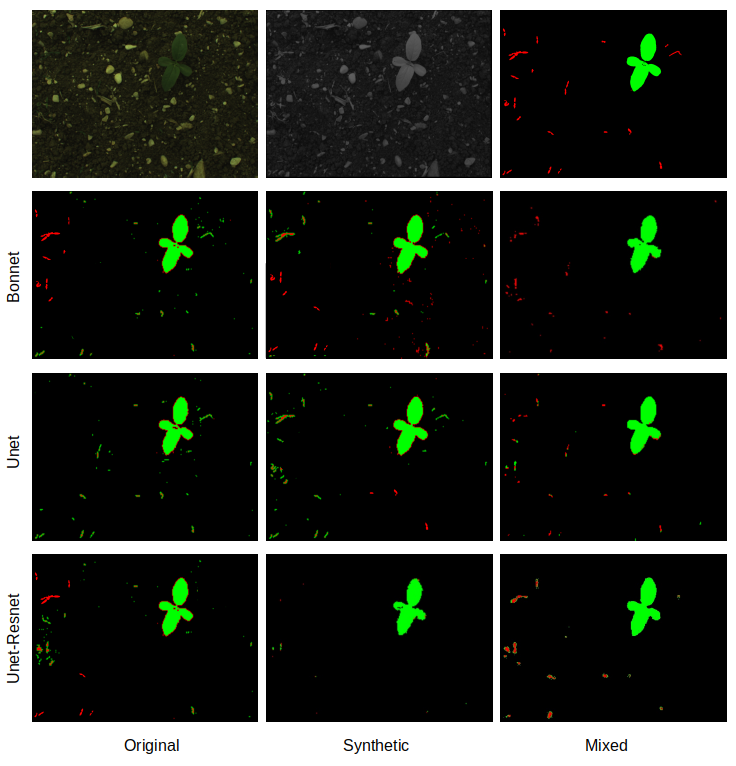}
    \caption{Examples of a segmented image from sunflower test set obtained by using three different segmentation networks trained with three different datasets. The first row of the image shows the input RGB, NIR images and its corresponding ground truth. The remaining rows show the segmentation results generated by the different networks on the $Original$, $Synthetic$, and $Mixed$ training datasets.}
    \label{fig:pred_ex_sunflower}
\end{figure}

\subsubsection{Experiment 4: Augmenting the Sunflower Dataset.}
 In this experiment, we used 500 images to train SPADE networks to generate sunflower crop, the second one trained to generate weeds (general types of weeds similar to those in Bonn dataset).
 In this experiment we used sunflower images.
 We created three different datasets:

 \begin{enumerate}
\item \textit{Original}, We took a total of 500 images from the  Sunflower dataset, randomly chosen among different days of acquisition in order to contain different growth-stages of the target crop. Then, we split it into a training set (350 images), and a test set (150 images). 
\item \textit{Synthetic Crop}, composed of 350 images with synthetic crop generated by using our architecture.
\item \textit{Mixed}, containing 350 images and composed by the union of 175 images from the Original dataset and 175 images with synthetic crops.
\end{enumerate}

We  used  the  three  datasets  described  above  to  train  three  state-of-the-art semantic  segmentation  networks, i.e., Bonnet, U-Net, and UNet-ResNet.
An example of segmentation results is shown in Fig. \ref{fig:pred_ex_sunflower}. Table \ref{tab:sunflower_segmentation_results} shows the quantitative results of the semantic segmentation on real images held out from sunflower dataset. In all cases, for testing we used 150 real images from Sunflower dataset not used during training. Also in the case of this dataset for all architectures the results show that the mIoU increases by using the original dataset augmented with the synthetic ones, as compared to using only the original dataset. Moreover, for this dataset architectures trained on synthetic data perform better than when trained on real data only.\\

All the datasets generated using the approach described in this work are publicly available and can be downloaded from:\\

\url{https://bit.ly/3hHenpE}

\subsection{Statistical Analysis}
To further support the significance of our results, a statistical analysis have been carried out. The null hypothesis, which we want to reject, says that there is no difference in the segmentation results if we use a dataset augmented with synthetic data or a dataset without. Since the null hypothesis is presumed to be true until the data shows enough evidence that it is not, we show here that a model built from a dataset augmented with synthetic data generates better results in the vast majority of the cases with respect to a model trained without them.

The analysis takes into account the Bonnet network because it was globally the best in our tests. We trained Bonnet on two different RGB sugar beet datasets, namely the \emph{Original} and \emph{Mixed} ones. We tested the two models, named $O$ and $M$ in Table \ref{tab:Statistical_results}, on 300 test images from the Bonn dataset, comparing the results on three evaluation metrics: $Accuracy$, Dice similarity coefficient ($DICE$), and intersection over union ($IoU$).
Although $Accuracy$ is easy to calculate and understand, it is not useful when the foreground and background classes are extremely imbalanced, i.e., when a class dominates the image and the other covers only a small portion of the image, which is the case in our scenario. Better metrics for dealing with the class imbalance issue are DICE and IoU.
The Dice score reflects both size and localization agreement, more in line with perceptual quality compared to pixel-wise accuracy \cite{Bertels2019}.


Results are shown in Table \ref{tab:Statistical_results}. Our method (model $M$) provides better results in 94.6\% of images using Accuracy, 96.3\% using DICE, and 96.6\% using IoU, with an average value of 95.3\%.
Since in about 95\% of the cases we have better results with synthetic augmentation, we can reject the null hypothesis.

\begin{table}[t]
\caption{Statistical analysis for the Bonnet architecture trained on two different sugar beet datasets and tested on $300$ images.}
\begin{center}
\begin{tabular}{ c  c | c }
 \hline
 Evaluation & \multicolumn{2}{c}{\# of images where}\\
metrics &  $M$ is better than $O$ & $O$ is better than $M$\\
\hline 
\textit{Accuracy}  
   
        & 284   
               & 16 
                  \\ \hline
                  
\textit{DICE}

         & 289   
               & 11   \\      \hline
\textit{IoU}  

         & 290 
               & 10 
              \\
                         \hline
       
        \end{tabular}
\end{center}
\label{tab:Statistical_results}
\end{table}

\subsection{Comparison with a Non-Conditional GAN}
\begin{figure}[t]
\centering
\begin{subfigure}{.5\textwidth}
  \centering
  \includegraphics[width=0.6\linewidth]{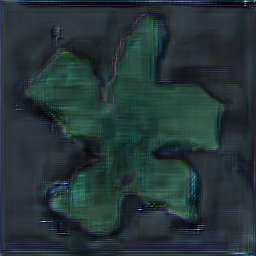}
  \caption{RGB. }
  \label{fig:sub1}
\end{subfigure}%
\begin{subfigure}{.5\textwidth}
  \centering
  \includegraphics[width=0.6\linewidth]{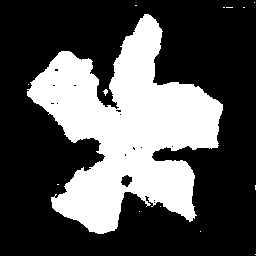}
  \caption{Mask.}
  \label{fig:sub2}
\end{subfigure}
\caption{Example of a sugar beet generated by a non-conditional GAN: (a) RGB image; (b) Segmentation mask obtained by thresholding the mask generated by the GAN.}
 \label{fig:-gan_output}
\end{figure}

The proposed system leverages a cGAN to generate data to be used to augment the available training data, using plant masks as an essential prerequisite. But what if we use a non-conditional generic Generative Adversarial Networks (GAN) \cite{GoodFellow2014GAN} modified to generate both the images and the segmentation masks? To answer this question, we designed a specific experiment in which a generic GAN has been modified and trained so to generate \emph{both} RGB images of crop  and soil \emph{and} the related segmentation masks (i.e., the plant and soil masks).\\
We exploited the GAN architecture presented in \cite{DCGAN}, modifying it in order to generate 4-channels images (i.e., the RGB image plus the segmentation mask). The segmentation masks of the training dataset defines one class (i.e., crop) with the value 1 and the other class (i.e., soil) with the value -1; we converted the (continuous) masks generated by the GAN simply by applying a threshold operator with threshold 0.
As conventional GANs, the generator network takes noise as input. Early experiments showed poor results in generating entire agricultural scenes (crop, weeds and soil) with accurate segmentation masks at full image resolution. In particular, the generated masks were qualitatively inadequate and inconsistent with real plants. We believe that this fact mostly derives from the limited size of the available training datasets (around 1,000 images in ours case). Hence, we decided to simplify the task, focusing on 256$\times$256 patches depicting single instance of crops, and soil as background. With this setup. the GAN results were apparently convincing (e.g., see Fig. \ref{fig:-gan_output}). The dataset used to train the GAN was composed of 2,000 patches with related segmentation masks taken from the Bonn sugar beet dataset.

\begin{table}[t]
\centering
\caption{Pixel-wise segmentation performance for Bonnet architecture, trained on three different datasets.}
\begin{tabular}{l c c c }
\hline
  \multirow{2}{*}{Model} &  &  \multicolumn{2}{c}{IoU}   \\ 
 \cline{3-4}
& mIoU & Soil  &
     Crop \\
\hline
\textit{Original} & 0.85 & 0.94 & 0.76 \\ 
\textit{Original+Ours} &  \textbf{0.94}&\textbf{0.98}&
     \textbf{0.89}   \\   
\textit{Original+GAN}& 0.61 & 0.88& 0.34  \\ 

\hline
\end{tabular}
\label{tab:DCGAN_evaluation}
\end{table}


We created three datasets:
\begin{itemize}
    \item \emph{Original}: 2,000 crop and soil patches extracted from the Bonn dataset
    \item \emph{Original + Ours}: \emph{Original} dataset augmented with 500 patches generated by our approach
    \item \emph{Original + Ours}: \emph{Original} dataset augmented with 500 patches generated by the non-conditianl GAN as previously described
\end{itemize}

We trained the Bonnet network on such datasets, testing the segmentation results on 300 test images from the Bonn dataset: the results are reported in Table  \ref{tab:DCGAN_evaluation}. Consistent with the previous results, our method allows to improve the performance compared to the \emph{Original} dataset (9\% increase in mIoU). Conversely, the model trained with the images generated by the GAN achieves poor performance. This result is mainly ascribable to the shape (i.e., the mask) and texture of the generated plants from the GAN, which do not comply to those of real plants.

%
                



\section{Conclusions} \label{sec:conclusions}
This paper introduces a data augmentation strategy that leverages a cGAN to generate entire agricultural scenes by synthesizing only the most relevant objects for segmentation purposes. The core of the proposed approach lies in exploiting the shapes of real objects to condition the trained generative models. The existing shapes are extracted from real-world labeled images. In addition, the generation process also synthesizes the NIR channel. The synthetically augmented dataset, obtained in this way, can then be used to train a semantic segmentation network. We applied this method to the crop/weed segmentation problem. As a further contribution, we also introduce and made publicly available with this paper a new crop/weed segmentation dataset, the Sunflower Dataset. Two kinds of quantitative evaluation have been carried out. In the first one, we test the cGAN generalization properties. Our experiments prove that with a small number of images we are able to generate good synthetic plant samples. 
The second evaluation aims to demonstrate that the cGAN augmented datasets can improve the performance of different state-of-the-art segmentation architectures. The results show that the segmentation quality increases by using the original dataset augmented with the synthetic ones, with respect to using only the original dataset. 
We believe that our method can serve as a valid tool for creating training frameworks for segmentation problems, allowing to improve segmentation performance, while reducing the amount of required labeled data.

\balance

%
%
\bibliographystyle{splncs03}
\bibliography{biblio}

\end{document}